\documentclass[default]{sn-jnl}

\usepackage{times}
\usepackage{graphicx}%
\usepackage{multirow}%
\usepackage{amsmath,amssymb,amsfonts}%
\usepackage{amsthm}%
\usepackage{mathrsfs}%
\usepackage[title]{appendix}%
\usepackage{xcolor}%
\usepackage{textcomp}%
\usepackage{manyfoot}%
\usepackage{booktabs}%
\usepackage{algorithm}%
\usepackage{algorithmicx}%
\usepackage{algpseudocode}%
\usepackage{listings}%
\usepackage{enumitem}
\usepackage{overpic}
\usepackage{adjustbox}
\usepackage{subcaption}
\usepackage{wrapfig}
\usepackage{pifont}

\usepackage{xspace}

\usepackage{mathrsfs}

\newcommand{\blue}[1]{\textbf{\textcolor{blue}{#1}}}
\newcommand{\red}[1]{\textbf{\textcolor{red}{#1}}}

\newcommand{\nameofmethod}{ControlSR}
\newcommand{\myPara}[1]{\vspace{5pt}\noindent\textbf{#1}}

\begin{document}

\title[Article Title]{\nameofmethod{}: Taming Diffusion Models for Consistent Real-World Image Super Resolution}


\author[1,2]{Yuhao Wan}\email{peaeswyh@gmail.com}
\equalcont{These authors contributed equally to this work.}

\author[2]{Peng-Tao Jiang}\email{
pt.jiang@vivo.com}
\equalcont{These authors contributed equally to this work.}

\author*[1]{Qibin Hou}\email{houqb@nankai.edu.cn}

\author[2]{Hao Zhang}

\author[2]{Jinwei Chen}

\author[1]{Ming-Ming Cheng}\email{cmm@nankai.edu.cn}

\author[2]{Bo Li}

\affil*[1]{VCIP, CS, Nankai Univeristy, Tianjin, 300350, China}

\affil[2]{vivo Mobile Communication Co., Ltd, Hangzhou, China}


\abstract{We present \nameofmethod{}, a new method that can tame Diffusion Models for consistent real-world image super-resolution (Real-ISR).
Previous Real-ISR models mostly focus on how to activate more generative priors of text-to-image diffusion models to make the output high-resolution (HR) images look better.
However, since these methods rely too much on the generative priors, the content of the output images is often inconsistent with the input LR ones.
To mitigate the above issue, in this work, we tame Diffusion Models by effectively utilizing LR information to impose stronger constraints on the control signals from ControlNet in the latent space.
We show that our method can produce higher-quality control signals, which enables the super-resolution results to be more consistent with the LR image and leads to clearer visual results.
In addition, we also propose an inference strategy that imposes constraints in the latent space using LR information, allowing for the simultaneous improvement of fidelity and generative ability.
Experiments demonstrate that our model can achieve better performance across multiple metrics on several test sets and generate more consistent SR results with LR images than existing methods.
Our code is available at \url{https://github.com/HVision-NKU/ControlSR}.}




\maketitle

\begin{figure*}[h]
    \centering
    \small
    \setlength{\abovecaptionskip}{0pt}
    \begin{overpic}[width=\linewidth]{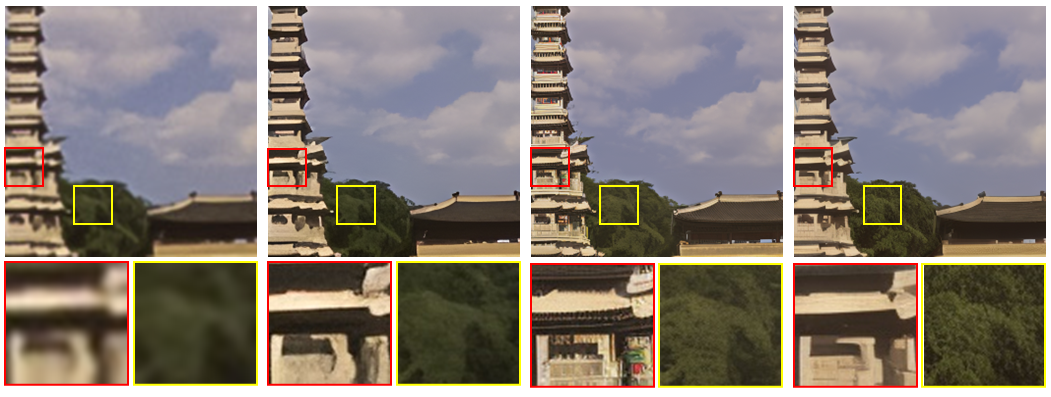}
    \put(2, -2){\small\textcolor{black}{(a) Zoomed LR Image}}
    \put(28, -2){\small\textcolor{black}{(b) Real-ESRGAN}}
    \put(57.5, -2){\small\textcolor{black}{(c) SeeSR}}
    \put(77, -2){\small\textcolor{black}{(d) ControlSR (Ours)}}
    \end{overpic}
    \vspace{8pt}
    \caption{Visual comparisons with recent state-of-the-art Real-ISR methods. Real-ESRGAN~\cite{Real-ESRGAN} results in a lack of generated details. SeeSR~\cite{SeeSR} uses semantic information to activate more generative priors of the SD model but results in \textbf{inconsistent} content with the LR image. Our results can properly generate details and have better visual effects.}
    \label{fig:first_fig}
    \vspace{2pt}
\end{figure*}
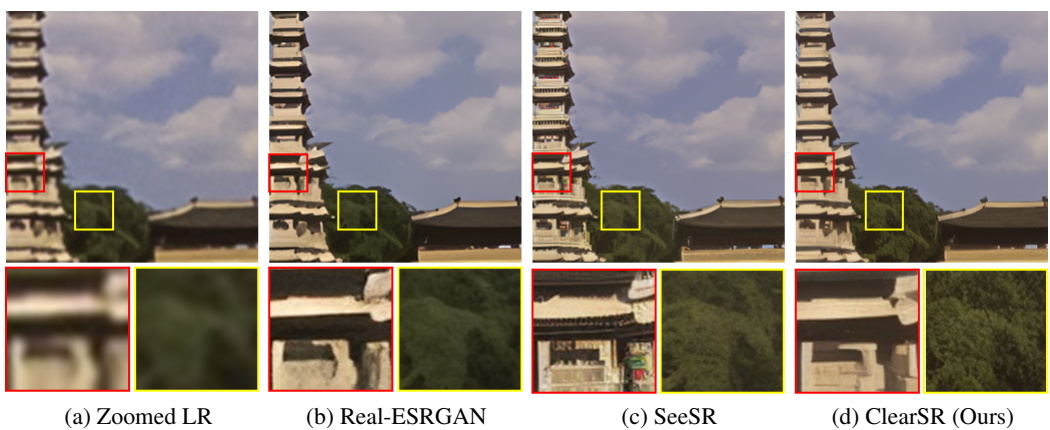

\section{Introduction}
\label{sec:introduction}
Real-world Image Super-Resolution (Real-ISR) aims to restore a high-resolution (HR) image from its low-resolution (LR) version in real-world scenarios. 
Unlike traditional Image Super-Resolution (ISR), Real-ISR requires modeling complex degradations in the real world, which further tests the models' capability of generating image details.
Some researchers~\cite{Real-ESRGAN, FeMaSR, BSRGAN} have used stacked convolutional blocks or transformer-based blocks to build models, or GANs to help generate details, achieving remarkable results. 
However, because of insufficient generative capability, these models are limited in generating fine details.

Recently, Diffusion Models (DMs) have achieved notable performance in various tasks. 
Specifically, the pre-trained text-to-image (T2I) models~\cite{Imagen, rombach2022high}, such as Stable Diffusion (SD), have a gift in powerful generative priors, which can help generate details needed for Real-ISR. 
Since then, many SD-based Real-ISR works~\cite{StableSR, DiffBIR, PASD, SeeSR, SUPIR, CoSeR} have emerged. 
However, pre-trained SD models are originally designed for image generation and directly using them for Real-ISR as done in previous work~\cite{StableSR} may lead to super-resolution results with inconsistent content with the input LR images because of the overly strong generative priors.
Therefore, \emph{how to tame SD models to avoid the generation of inconsistent content has become a challenge on this topic.}
This requires a strong control ability for diffusion models to generate desired region details.

A common approach to mitigate the above issue in previous work~\cite{DiffBIR} is to use diffusion adapters, such as ControlNet~\cite{zhang2023adding}, to process the LR image. 
To take advantage of the adapters, PASD~\cite{PASD} introduces additional cross-attention layers to integrate the control signals produced by ControlNet into the UNet, demonstrating better consistency between the output and the input LR images.
However, this method mainly focuses on the utilization of the control signals but does not consider the way of constructing them with high quality.
Therefore, while this method is beneficial for consistency, it struggles to ensure the generation of fine details.
To improve the detail information generation ability, some recent works~\cite{SeeSR, CoSeR} propose to extract semantic information from the LR image to activate more generative priors. 
Due to the introduction of too many semantic information, these methods can produce visually pleasing images but often result in too much content that is inconsistent with the input image, as shown in Figure~\ref{fig:first_fig}.

\begin{figure*}[t]
    \centering
    \small
    \setlength{\abovecaptionskip}{0pt}
    \begin{overpic}[width=\linewidth]{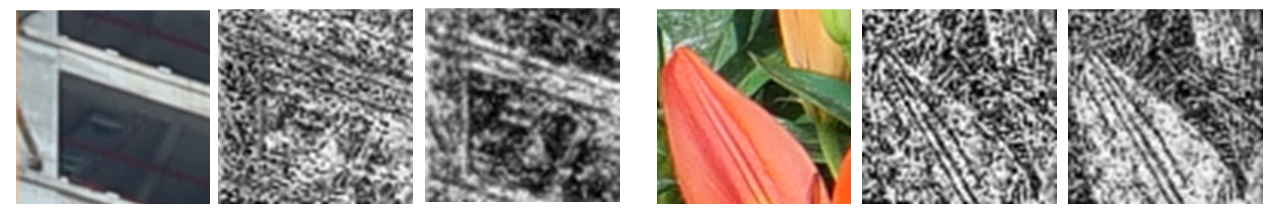}
    \put(7.5, -2.7){\small\textcolor{black}{LR}}
    \put(19.5, -1.8){\small\textcolor{black}{ControlNet}}
    \put(18.5, -4.0){\small\textcolor{black}{$D_{kl}=0.055$}}
    \put(38.5, -1.8){\small\textcolor{black}{Ours}}
    \put(34.5, -4.0){\small\textcolor{black}{$D_{kl}=0.004$}}
    
    \put(57.5, -2.7){\small\textcolor{black}{LR}}
    \put(69.5, -1.8){\small\textcolor{black}{ControlNet}}
    \put(68.5, -4.0){\small\textcolor{black}{$D_{kl}=0.107$}}
    \put(88.5, -1.8){\small\textcolor{black}{Ours}}
    \put(84.5, -4.0){\small\textcolor{black}{$D_{kl}=0.042$}}
    \end{overpic}
    \vspace{15pt}
    \caption{Analysis of the role of the latent LR embeddings constraint. $D_{kl}$ represents the KL divergence between the control signals and latent LR embeddings. We visualize the control signals with PCA~\cite{PCA}. One can observe that the control signals of ControlNet have higher $D_{kl}$ and cannot preserve the LR information well. However, our results have lower $D_{kl}$ and have sharper outlines, indicating that our model can extract LR information better. Further analysis can be seen in Section~\ref{sec:further}.}
    \label{fig:pca}
\end{figure*}
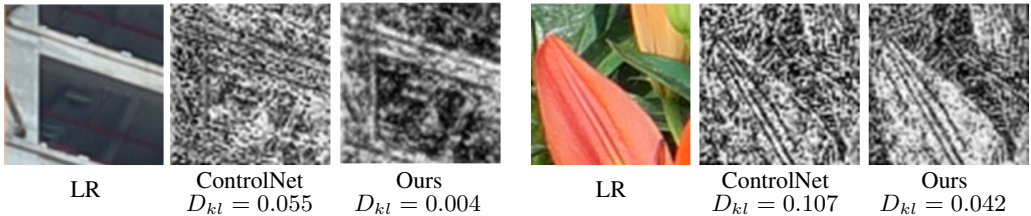

The above methods have shown that semantic information can be used as conditions to adjust the control signals and the quality of the visual results can be largely improved. 
However, the problem is that the semantic information is more abstract than the LR image itself, which can lead to content inconsistency in the generated results.
To figure out how to better take advantage of the control signals, we first attempt to visualize them from ControlNet in Figure~\ref{fig:pca}.
The visualization results show that these control signals cannot actually preserve the LR information well.
Considering the coarseness of the semantic information, we propose imposing stronger constraints on the control signals using the latent LR embeddings.
These LR embeddings can be produced by the pre-trained VAE encoder, which retains rich LR information. 
Since the control signals provided by the original ControlNet exhibit a significant loss of LR information, these constraints can be effectively used to guide the control signals.

To be specific, we take advantage of the latent LR embeddings by designing two new modules, called Detail Preserving Module (DPM) and Global Structure Preserving Module (GSPM).
Their goals are to embed the latent LR embeddings through window-based cross-attention into different layers of ControlNet to enhance image details, and meanwhile preserve the structural information of the LR images, respectively.

Moreover, we show that the use of latent LR embeddings in the inference stage is also able to address the limitation of previous methods that could only enhance fidelity while not improving generative capability. 
We achieve this by introducing the Latent Space Adjustment (LSA) strategy. 
This strategy uses latent LR embeddings to adjust the latent space at both earlier and later timesteps, allowing for a wide range of adjustments to the super-resolution results (over 2dB in PSNR and 0.1 in MANIQA). 
With appropriate settings, both the fidelity and generative capability of our model can be enhanced.

To show the advantages of the proposed \nameofmethod{}, we compare with a series of recent state-of-the-art SR models based on diffusion on widely-used datasets, such as DRealSR and RealSR.
Extensive experiments demonstrate that our \nameofmethod{} has superior generation capabilities and can produce higher-quality super-resolution results. As shown in Figure~\ref{fig:first_fig}, one can observe that our results can properly generate details and have better visual effects.
Our contributions can be briefly summarized as follows:
\begin{itemize}[leftmargin=*]
\item We present \nameofmethod{}, a new method that can tame Diffusion Models for consistent real-world image super-resolution (Real-ISR). The cores are the DPM and GSPM that can utilize the latent LR embeddings properly to impose stronger constraints on the control signals.
\item  We show that the latent LR embeddings can be used to adjust the latent space during the inference stage, which brings improvement of the fidelity and generation ability simultaneously.
\item Our proposed \nameofmethod{} outperforms previous models on multiple metrics on different test sets. The super-resolution results generated by \nameofmethod{} contain rich generated details and meanwhile show better consistency with LR images.
\end{itemize}

\section{Related work}

\subsection{Image Super-Resolution}
Image Super-Resolution (ISR) aims to restore a high-resolution (HR) image from its low-resolution (LR) version. Traditional ISR works are usually based on stacked CNN or transformer layers and are learned under a known degradation.
Since SRCNN~\cite{SRCNN} introduced CNN into the field of image super-resolution and achieved better results than traditional methods, many excellent works have emerged~\cite{SRCNN, RCAN, SAN, HAN, SRDenseNet, VDSR, RDN, shi2016real, CARN, EDSR, NLSA, FSRCNN, LAPAR}. These works are mainly designed based on stacked CNNs.
After that, some researchers applied Swin Transformer~\cite{SwinTrans} to the image super-resolution task and achieved impressive success~\cite{SwinIR, HAT, SRFormer, DAT, ELAN, ATD}.
In these works, such as SRFormer~\cite{SRFormer} and HAT~\cite{HAT}, improvements were made to the window attention mechanism to enable the model to expand the receptive field. HAT~\cite{HAT} and DAT~\cite{DAT} attempted to introduce channel attention. ATD~\cite{ATD} grouped tokens at the spatial level, making the attention mechanism better suited to the requirements of SR tasks. However, as the degradation is usually simple and known, the application scope of this task is limited. 
In recent years, attention has shifted toward more practically valuable topics, such as Real-world Image Super-Resolution~\cite{BSRGAN, DASR, Real-ESRGAN, StableSR, DiffBIR, SeeSR, PASD, Desra}.

\subsection{Real-World Image Super-Resolution}
Real-world Image Super-Resolution (Real-ISR) has become a popular topic in recent years. 
Compared to traditional ISR, Real-ISR requires modeling complex degradations in the real world, which further tests the generative capabilities of models and offers greater practical values.
Many studies have used GANs~\cite{Real-ESRGAN, BSRGAN, LDL} for Real-ISR tasks due to its excellent detail generation capabilities, demonstrating competitive results~\cite{BSRGAN, Real-ESRGAN, LDL, DASR}. However, GAN-based methods often produce unnatural artifacts, limiting their applications in Real-ISR tasks.
Recently, since the introduction of DDPM~\cite{DDPM}, Diffusion Models (DMs) have secured a significant position in the field of image synthesis. After some exploration~\cite{DPM-solver, FastDPM, san2021noise}, ~\cite{rombach2022high} reduced the computational cost of DMs, broadening its application range. 

Due to the outstanding success of DMs in various computer vision tasks, some researchers have begun to use them for Real-ISR tasks~\cite{ResShift, SinSR, DiffIR}, but the generative capabilities of these models are still limited.
As the pre-trained text-to-image (T2I) DMs, such as Stable Diffusion (SD) have powerful generative priors, which can help generate details needed for Real-ISR. StableSR~\cite{StableSR} has used SD for the first time to conduct Real-ISR tasks and demonstrates impressive detail generation capabilities.
However, the overly strong generative ability of the pre-trained T2I models often leads to inconsistent super-resolution results. Therefore, how to tame SD models to avoid the generation of inconsistent content has become a challenge on this topic. DiffBIR~\cite{DiffBIR} has used ControlNet~\cite{zhang2023adding} to provide appropriate control signals for SD, improving the generation effect of the model. 
On this basis, PASD~\cite{PASD} focuses on the control signals provided by ControlNet, making more efficient use of them. 
Our work also focuses on how to tame SD models. By analyzing previous methods~\cite{PASD, SeeSR}, we find PASD~\cite{PASD} did not improve the control signals themselves, and the usage of semantic information~~\cite{SeeSR} is coarse and leads to inconsistent SR results.
As a result, we focus on latent LR embeddings provided by the pre-trained VAE encoder and use it to optimize the control signals at both detail and structure levels. 

\begin{figure*}[t]
    \centering
    \small
    \setlength{\abovecaptionskip}{0pt}
    \includegraphics[width=\linewidth]{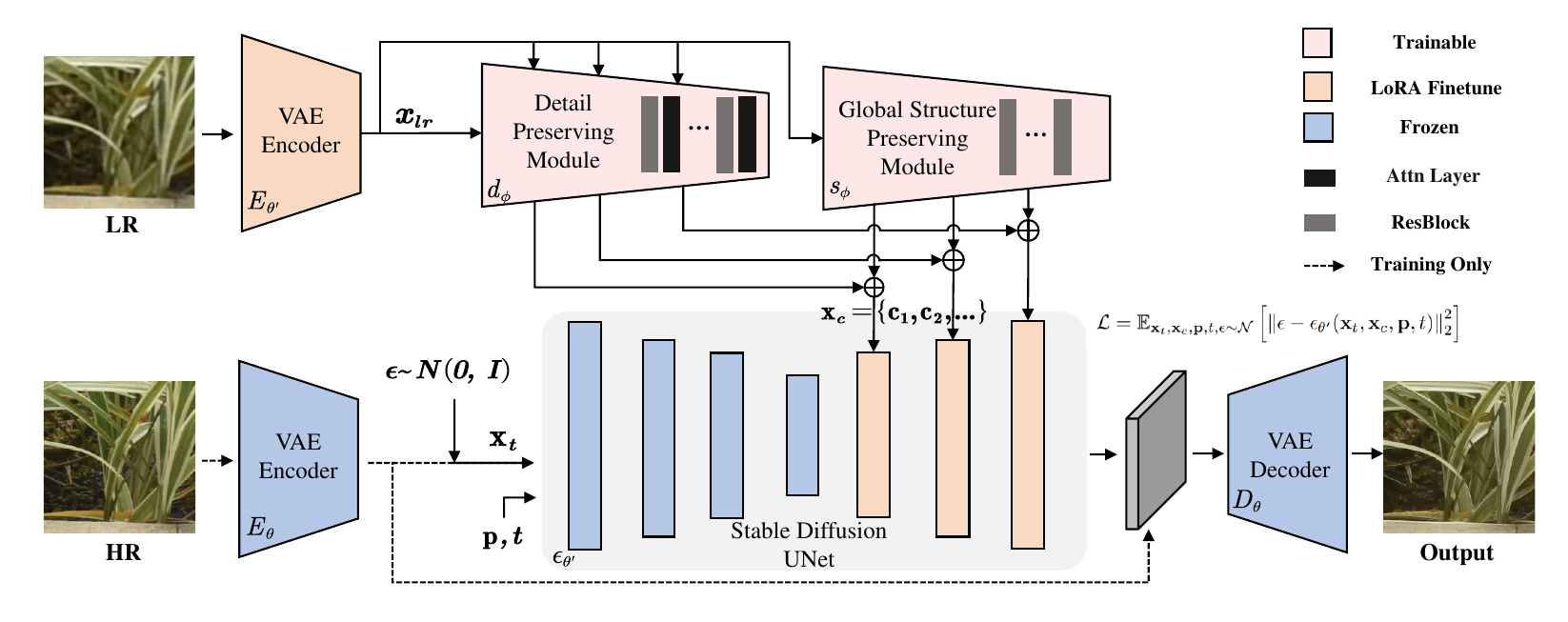}
    \caption{Overview of our \nameofmethod{}. Our \nameofmethod{} consists of the pre-trained Stable Diffusion (SD), the Detail Preserving Module (DPM), and the Global Structure Preserving Module (GSPM). To produce high-quality control signals, we let the LR image pass through the LoRA finetuned VAE Encoder first to obtain latent LR embeddings $\mathbf{x}_{lr}$. Then, we collect the control signals $\mathbf{x}_c = \{\mathbf{c}_1, \mathbf{c}_2, \dots\}$ by inputting $\mathbf{x}_{lr}$ into the DPM and the GSPM and summing their outputs. We feed the control signals into the decoder of SD UNet to control the HR image generation.}
    \label{fig:model}
\end{figure*}

\section{Methodology}

\subsection{Overall Architecture of \nameofmethod}

Our intention is to model high-quality control signals to better tame SD models to generate more consistent HR images.
We achieve this by presenting two new modules, named Detail Preserving Module (DPM) and Global Structure Preserving Module (GSPM), which integrate fine details and structural information respectively from LR images into the control signal.
Figure~\ref{fig:model} shows the overall pipeline of our \nameofmethod{}.

\begin{figure*}[h]
    \centering
    \small
    \setlength{\abovecaptionskip}{0pt}
    \includegraphics[width=\linewidth]{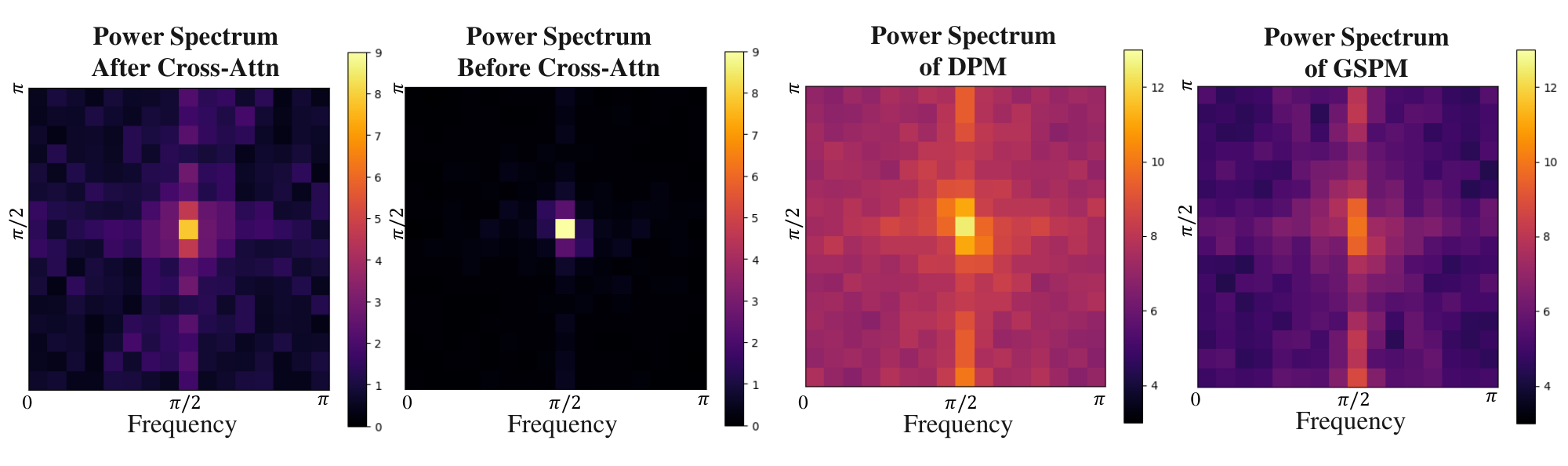}
    \caption{Power spectrum visualization of the intermediate features. The two images on the left show that the cross-attention layer can increase high-frequency information, and the two images on the right show that DPM contains more high-frequency information than GSPM.}
    \label{fig:method_vis}
\end{figure*}

During the training process, the objective of the Diffusion Model (DM) is to learn the probability distribution of the reverse denoising process. Specifically, denote the LoRA finetuned SD UNet, the LoRA finetuned VAE encoder, the pre-trained VAE encoder, and the pre-trained VAE decoder as $\epsilon_{\theta'}$, $E_{\theta'}$, $E_{\theta}$, and $D_{\theta}$, respectively. 
Denote DPM and GSPM as $d_{\phi}$ and $s_{\phi}$. 
For a randomly sampled time step $t$ and a high-quality image $\mathbf{I}_{hq}$, 
let $\mathbf{I}_{hq}$ pass through $E_{\theta}$ and perform the noise addition process to obtain $\mathbf{x}_{t}$. 
Sending the low-quality image $\mathbf{I}_{lr}$ into $E_{\theta'}$ yields the latent LR embeddings $\mathbf{x}_{lr}$. 
Then, we can collect the control signals $\mathbf{x}_c = \{\mathbf{c}_1, \mathbf{c}_2, \dots\}$ by inputting $\mathbf{x}_{lr}$ into $d_{\phi}$ and $s_{\phi}$ and summing their outputs. 
Similar to PASD~\cite{PASD} and CoSeR~\cite{CoSeR}, we let the LR image pass through the CLIP image encoder to obtain the image-level feature $\mathbf{p}$ and replace the null-text prompt in the UNet decoder. The optimization objective can be formulated as:
\begin{equation}
\mathcal{L}=\mathbb{E}_{\mathbf{x}_{t}, \mathbf{x}_{c}, \mathbf{p}, t, \epsilon\sim\mathcal{N}}\left[\left\| \epsilon - \epsilon_{\theta'}(\mathbf{x}_t, \mathbf{x}_c, \mathbf{p}, t) \right\|_2^2\right],
\end{equation}
where the $\epsilon$ is the added noise.

As mentioned in a previous work~\cite{SUPIR}, the pre-trained VAE encoder is unsuitable for encoding LR images, because it was not trained on LR images. 
During training, unlike previous works, such as SUPIR~\cite{SUPIR} and SeeSR~\cite{SeeSR}, that introduce a new loss or design a new encoder, we simply add LoRA layers to the pre-trained VAE encoder to tackle this issue. Therefore, there is no need to separately train an encoder or design a new encoder. We also add LoRA layers to the SD UNet decoder to adapt the model to the mixed control signals.

\subsection{High-Quality Control Signal Modeling}
As mentioned above, we intend to utilize LR information to impose stronger constraints on the control signals.
We achieve this by adjusting the control signals at both the detail and structure levels, which corresponds to two new modules, called Detail Preserving Module (DPM) and Global Structure Preserving Module (GSPM), respectively.
In what follows, we will give their detailed descriptions.

\myPara{Detail Preserving Module.}
Our DPM aims to constrain ControlNet at the detail level. As ControlNet contains generative priors of SD, the detailed information of latent LR embeddings cannot be preserved well (See Figure~\ref{fig:pca}). 
As a result, we use window-based cross-attention layers to integrate the latent LR embeddings into different layers of ControlNet. These cross-attention layers are placed after text cross-attention layers.
Specifically, let the newly added cross-attention layer be denoted as $\operatorname{CA}$. Given the intermediate feature $\mathbf{x}_{d} \in \mathbb{R}^{L \times C} $, we let it pass 
through the linear layer and window partition yields $\mathbf{Q} \in \mathbb{R}^{N \times S^2 \times C }$. 
Then, let the latent LR embeddings $\mathbf{x}_{lr} \in \mathbb{R}^{l \times c }$ pass through a linear layer and window partition, yielding $\mathbf{K} \in \mathbb{R}^{N \times s^2 \times C} $ and 
$\mathbf{V} \in \mathbb{R}^{N \times s^2 \times C} $, respectively. 
Here, $N$ is the number of windows, $S$ is the side length of each window of $\mathbf{Q}$, 
$s$ is the side length of each window of $\mathbf{K}$ and $\mathbf{V}$, 
$L$ and $C$ are the token number and channel number of $\mathbf{x}_{d}$, and $l$ as well as $c$ are the token number and channel number of $\mathbf{x}_{lr}$. 
The formulation can be written as follows:
\begin{equation}
\operatorname{CA}(\mathbf{Q}, \mathbf{K}, \mathbf{V})= 
\operatorname{Softmax}\left(\frac{\mathbf{Q} \mathbf{K}^{T}}{\sqrt{d_{k}}} + \mathbf{B} \right) \mathbf{V},
\end{equation}
where $\mathbf{B}$ is an aligned relative position embedding and $\sqrt{d_{k}}$ is a scaling factor 
as defined in ~\cite{dosovitskiy2020image}.

\myPara{Global Structure Preserving Module.}
Our GSPM aims to constrain ControlNet at the structure level. 
GSPM is an independent module that removes the transformer blocks and only retains the ResBlocks from the ControlNet.
Since the attention layer is based on weighted calculations between features, it may ignore the original spatial structure.
Thus, excluding the attention layers can preserve the structural information~\cite{yu2024convolutions} that helps generate a consistent HR image with the input LR one.
GSPM can present multi-scale control signals consistent in shape with DPM. 
We sum up the two to form the final control signals $x_c = \{c_1, c_2, \dots\}$.

\myPara{Analysis.}
We demonstrate the effectiveness of adding constraints to ControlNet first. As shown in Figure~\ref{fig:pca}, we evaluate the deviation between the control signal and the LR information by calculating the KL divergence $D_{kl}$ between the control signals and the latent LR embeddings. Compared to the model that only uses ControlNet (also trained), the control signal output of our model exhibits a lower $D_{kl}$, indicating that latent LR embeddings successfully constrain ControlNet. Furthermore, we visualized the control signal using PCA~\cite{PCA}, which reveals that our control signal maintains the LR information effectively, demonstrating that our method can model high-quality control signals.

Next, we briefly analyze why our DPM and GSPM help. In Figure~\ref{fig:method_vis}, we use the power spectrum of intermediate features to validate the effectiveness of our DPM and GSPM. The two images on the left show the power spectrum of features in the DPM before and after passing through one cross-attention layer. We can see that after the cross-attention layer, the intermediate features contain more high-frequency components, indicating that more detailed information has been extracted, which aligns with our design intent. The two images on the right show the power spectrum of the control signals from DPM and GSPM. It can be seen that the output from DPM contains more high-frequency information which is helpful for reconstructing details while GSPM mainly contains low-frequency information which preserves structural information.

\subsection{Latent Space Adjustment strategy in inference stage}
Previous work has pointed out that adding additional LR information during the inference stage can help improve fidelity~\cite{SUPIR,SeeSR}. 
However, the improvement in fidelity comes at the expense of reducing the generative capability of the models.
This type of unidirectional adjustment strategy has a negative impact on the image details after super-resolution, affecting the visual effect.
Unlike previous works, we propose the Latent Space Adjustment (LSA) strategy, which can improve either fidelity or generation. 
Moreover, our strategy can improve the fidelity and generation simultaneously through appropriate settings.

\begin{figure*}[]
    \centering
    \small
    \setlength{\abovecaptionskip}{0pt}
    \includegraphics[width=\linewidth]{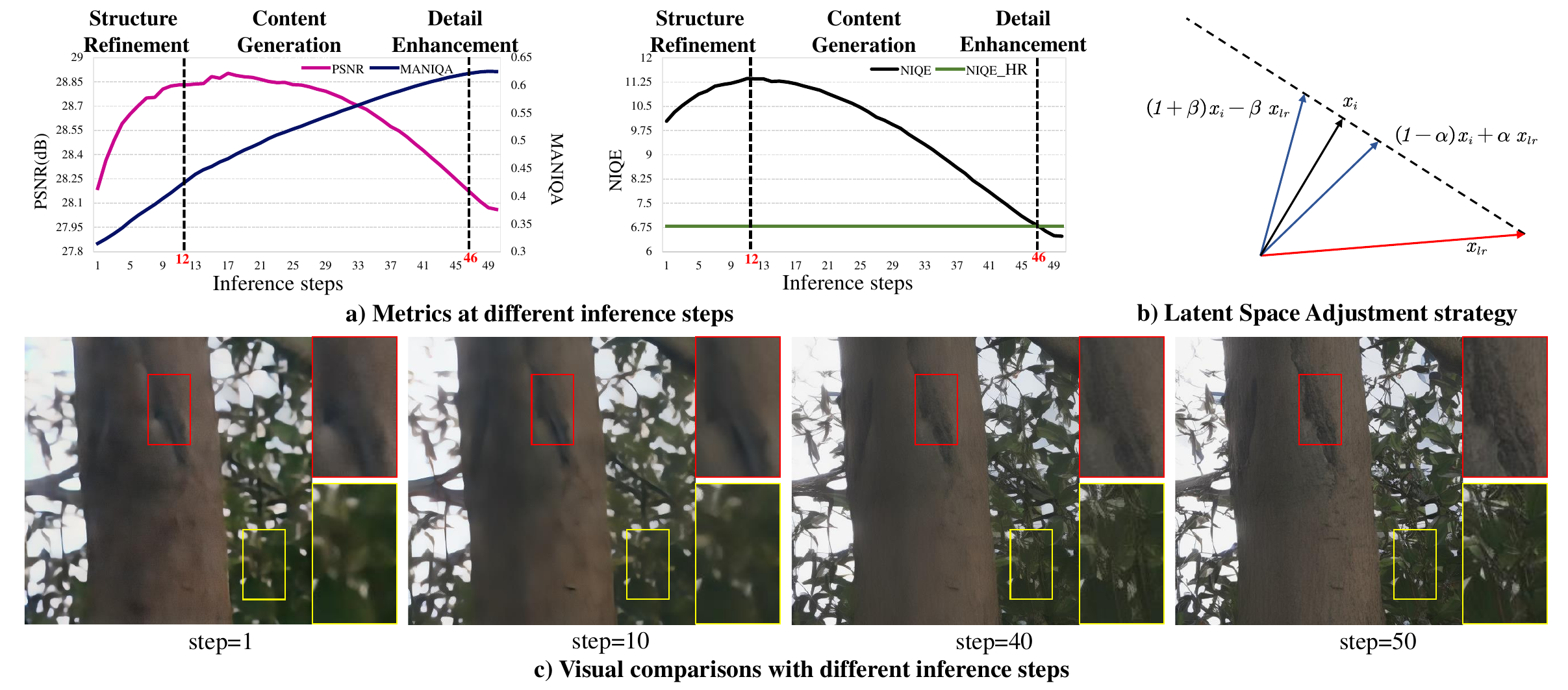}
    \caption{Overview of our Latent Space Adjustment strategy. a) shows the average PSNR, MANIQA, and NIQE curves of the DRealSR test set. b) demonstrates our Latent Space Adjustment strategy. c) shows the images at different steps.}
    \label{fig:lsa}
\end{figure*}
As shown in Figure~\ref{fig:lsa}(a), during the inference stage, the PSNR score increases first and then decreases as the number of steps increases. 
This is because the model performs structural refinement in the early steps while generation in the later steps~\cite{CCSR}. 
For the middle steps of the inference stage, the model focuses on content generation.
As shown in Figure~\ref{fig:lsa}(c), at around 40$_{th}$ step, the model can already determine most of the information in the image, but it is difficult to generate realistic textures. 
This indicates that detail enhancement is mainly in the last few steps.
This motivates us to divide the whole inference stage into three parts: structure refinement, content generation, and detail enhancement.

Based on the analysis above, we propose the Latent Space Adjustment  (LSA) strategy. 
We notice that an inherent property of LR images is that it mainly contains structural information and has less details compared to HR images.
We take advantage of this property and move the output of each inference step away from the latent LR embeddings in the latent space in the later steps of the inference stage so that the model can focus more on generating details~\cite{DiffBIR}.
In contrast, in the early steps, we let the output close to the latent LR embeddings, similar to previous work, to enhance the fidelity. 
As shown in Figure~\ref{fig:lsa}(a), we statistically select the highest point of the NIQE curve and the point where NIQE starts to fall below the HR image to split the inference stage into three parts.
The LR adjustments in the structure refinement and detail enhancement are referred to as Early-step LR Adjustment (ELA) and Later-step LR Adjustment (LLA), respectively. 
We use two factors $\alpha$ and $\beta$ to determine the control level. 
Figure~\ref{fig:lsa}(b) shows our LSA strategy. The $x_{i}$ is the predicted latent embeddings of $i$-th step. The formulation can be written as follows:
\begin{equation}
\operatorname{ELA}(x_i)=(1-{\alpha}) x_i + {\alpha}x_{lr}, \end{equation}
\begin{equation}\operatorname{LLA}(x_i)=(1+{\beta}) x_i - {\beta}x_{lr}.
\end{equation}
The experimental results show that using ELA can improve the fidelity of the model, and using LLA can improve the generation, solving the problem that previous methods~\cite{SeeSR, SUPIR} can only adjust in one direction. Moreover, the fidelity and generation of the model can be improved simultaneously through appropriate $\alpha$ and $\beta$ settings. (See Section~\ref{sec:ablation} for more discussions.)

\begin{table*}[!h]
\setlength{\abovecaptionskip}{2pt}
\caption{
Quantitative comparison of our \nameofmethod{} with recent state-of-the-art \textbf{Real-ISR} methods on five benchmark datasets.
The best performance is marked in \red{red} and the second best is marked in \blue{blue}. We compare \nameofmethod{*} with GAN-based and Diffusion-based methods (no generative priors), and \nameofmethod{} with SD-based methods. \nameofmethod{*} has the same structure as \nameofmethod{}, but with improved fidelity by modifying the LSA settings.
}
\label{tab:performance_sd+diffusion}
\setlength\tabcolsep{3pt}
\renewcommand{\arraystretch}{1.}
\small
\centering
\scalebox{0.9}{
\begin{tabular}{clccccccc}
\toprule
{Datasets}\hspace{3pt}  &  {Method} & {PSNR$\uparrow$} & {SSIM$\uparrow$} & {LPIPS$\downarrow$} & {NIQE$\downarrow$} & {MUSIQ$\uparrow$} & {MANIQA$\uparrow$} & {CLIPIQA$\uparrow$}  \\
\midrule 
\multirow{11}{*}{{DRealSR}} \hspace{3pt}

& \textbf{Real-ESRGAN}~\cite{Real-ESRGAN} & \blue{28.64} & \blue{0.8053} & \blue{0.2847} & \red{6.6928} & 54.18 & 0.4907 &  0.4422 \\
& \textbf{LDL}~\cite{LDL} & 28.21 & \red{0.8126} & \red{0.2815} & 7.1298 &  53.85 & \blue{0.4914} & 0.4310 \\

& \textbf{ResShift}~\cite{ResShift} & 28.46 & 0.7673 &  0.4006 & 8.1249 & 50.60 & 0.4586 & 0.5342 \\
& \textbf{SinSR}~\cite{SinSR} & 28.36 & 0.7515 & 0.3665 & 6.9907 & \blue{55.33} & 0.4884 & \blue{0.6383}\\
& \textbf{\nameofmethod{*(ours)}} & \red{29.00} & 0.7781 & 0.3281 & \blue{6.9796} & \red{62.74} & \red{0.5878} & \red{0.6585}  \\
\cmidrule(lr){2-9}

& \textbf{StableSR}~\cite{StableSR} & 28.03 &  0.7536 &  \blue{0.3284} & 6.5239 & 58.51 & 0.5601 & 0.6356 \\
& \textbf{PASD}~\cite{PASD} & 27.36 & 0.7073 & 0.3760 & \red{5.5474} & 64.87 & \blue{0.6169} & \blue{0.6808} \\
& \textbf{DiffBIR}~\cite{DiffBIR} & 26.71 & 0.6571 & 0.4557 & 6.3124 & 61.07 & 0.5930 & 0.6395\\
& \textbf{SeeSR}~\cite{SeeSR} & \blue{28.17} & \red{0.7691} & \red{0.3189} & 6.3967 & \blue{64.93} & 0.6042 & 0.6804  \\

& \textbf{\nameofmethod{(ours)}} & \red{28.22} & \blue{0.7538} & 0.3473 & \blue{6.0867} & \red{66.27} & \red{0.6246} & \red{0.6976} \\

\midrule

\multirow{11}{*}{{RealSR}} \hspace{3pt}

& \textbf{Real-ESRGAN}~\cite{Real-ESRGAN} & 25.69 & \red{0.7616} & \red{0.2727} & \blue{5.8295} & 60.18 & \blue{0.5487} &  0.4449 \\
& \textbf{LDL}~\cite{LDL} & 25.28 & \blue{0.7567} & \blue{0.2766} & 6.0024 &  \blue{60.82} & 0.5485 & 0.4477  \\
& \textbf{ResShift}~\cite{ResShift} & \red{26.31} & 0.7421 &  0.3460 & 7.2635 &  58.43 &  0.5285 & 0.5444  \\
& \textbf{SinSR}~\cite{SinSR} & \blue{26.28} & 0.7347 & 0.3188 & 6.2872 &  60.80 & 0.5385 & \blue{0.6122}  \\
& \textbf{\nameofmethod{*(ours)}} & 25.86 & 0.7141 & 0.3148 & \red{5.6775} & \red{67.70} & \red{0.6262} & \red{0.6560}  \\
\cmidrule(lr){2-9}

& \textbf{StableSR}~\cite{StableSR} & 24.70 & \blue{0.7085}&  \blue{0.3018} & 5.9122 &  65.78 & 0.6221 & 0.6178 \\
& \textbf{PASD}~\cite{PASD} & \blue{25.21} &  0.6798 &  0.3380 & 5.4137 & 68.75 & \blue{0.6487} & \blue{0.6620} \\
& \textbf{DiffBIR}~\cite{DiffBIR} & 24.75 & 0.6567 &  0.3636 &  5.5346 &  64.98 &  0.6246 &  0.6463 \\
& \textbf{SeeSR}~\cite{SeeSR} & 25.18 & \red{0.7216} & \red{0.3009} & \blue{5.4081} & \blue{69.77} & 0.6442 & 0.6612  \\
& \textbf{\nameofmethod{(ours)}} & \red{25.30} & 0.6911 & 0.3318 & \red{5.0642} & \red{69.83} & \red{0.6499} & \red{0.6960}  \\

\midrule

\multirow{11}{*}{{DIV2K-Val}} \hspace{3pt}

& \textbf{Real-ESRGAN}~\cite{Real-ESRGAN} & 24.29 & \red{0.6371} & \red{0.3112} & \red{4.6786} & 61.06 & \blue{0.5501} &  0.5277 \\
& \textbf{LDL}~\cite{LDL} & 23.83 & \blue{0.6344} & 0.3256 & \blue{4.8554} &  60.04 & 0.5350 & 0.5180 \\
& \textbf{ResShift}~\cite{ResShift} & \red{24.65} & 0.6181 &  0.3349 & 6.8212 &   61.09 &  0.5454 &  0.6071 \\
& \textbf{SinSR}~\cite{SinSR} & \blue{24.41} & 0.6018 &  \blue{0.3240} & 6.0159 &  \blue{62.82} &  0.5386 & \blue{0.6471} \\
& \textbf{\nameofmethod{*(ours)}} & 24.23 & 0.5958 & 0.3441 & 5.0315 & \red{66.90} & \red{0.6118} & \red{0.6788}\\ \cmidrule(lr){2-9}

& \textbf{StableSR}~\cite{StableSR} & 23.26 &  0.5726 &  \red{0.3113} & 4.7581 &  65.92 &  0.6192 & 0.6771\\
& \textbf{PASD}~\cite{PASD} & 23.14 &  0.5505 &   0.3571 & \red{4.3617} & \blue{68.95} & \red{0.6483} & 0.6788\\
& \textbf{DiffBIR}~\cite{DiffBIR} & 23.64 & 0.5647 &  0.3524 &  4.7042 &  65.81 &  0.6210 &  0.6704 \\
& \textbf{SeeSR}~\cite{SeeSR} & \blue{23.68} & \red{0.6043} & \blue{0.3194} & 4.8102 & 68.67 & 0.6240 & \blue{0.6936}  \\

& \textbf{\nameofmethod{(ours)}} & \red{23.86} & \blue{0.5796} & 0.3493 & \blue{4.6146} & \red{69.34} & \blue{0.6328} & \red{0.7040}\\
\bottomrule 
\end{tabular}
}
\end{table*}

\section{Experiments}
\subsection{Experiment Settings}
\label{sec:exp settings}

Following previous works~\cite{SeeSR, PASD}, for training, we train \nameofmethod{} on DIV2K~\cite{DIV2K}, Flickr2K~\cite{Flickr2K}, DIV8K~\cite{DIV8K}, OST~\cite{OST}, and the first 10K face images from FFHQ~\cite{FFHQ}. We use the degradation pipeline of Real-ESRGAN~\cite{Real-ESRGAN} to obtain LR/HR pairs.
For testing, we test \nameofmethod{} on DRealSR~\cite{DRealSR}, RealSR~\cite{RealSR}, and DIV2K-Val~\cite{DIV2K} using the same configurations as ~\cite{SeeSR}.
For evaluation, we employ 7 widely used metrics, including PSNR, SSIM, LPIPS, NIQE, MUSIQ, MANIQA, and CLIPIQA. We use PSNR, SSIM (calculated on the Y channel from the YCbCr space) to evaluate fidelity, LPIPS to evaluate perceptual quality, and NIQE, MUSIQ, MANIQA, and CLIPIQA to evaluate the generation ability of the model.

For implementation details, we use SD 2.1-base as our pre-trained T2I model. We use the Adam optimizer to train \nameofmethod{}. The total iteration, batch size, learning rate, inference step are set to 150K, 8, $5 \times 10^{-5}$, and 50, respectively. $\alpha$ and $\beta$ are set to 0.01 and 0.01 for \nameofmethod{} and 0.03 and 0.01 for \nameofmethod{*}, respectively. We also use the LR embeddings proposed by~\cite{SeeSR} in the inference stage to improve the fidelity. The training process is conducted on $512 \times 512$ resolution with 4 NVIDIA A40 GPUs. For inference, we use a spaced DDPM sampling schedule~\cite{IDDPM}.

\subsection{Further Analysis on LR latent embeddings constraint}
\label{sec:further}
\begin{figure*}[h]
    \centering
    \small
    \setlength{\abovecaptionskip}{0pt}
    \includegraphics[width=\linewidth]{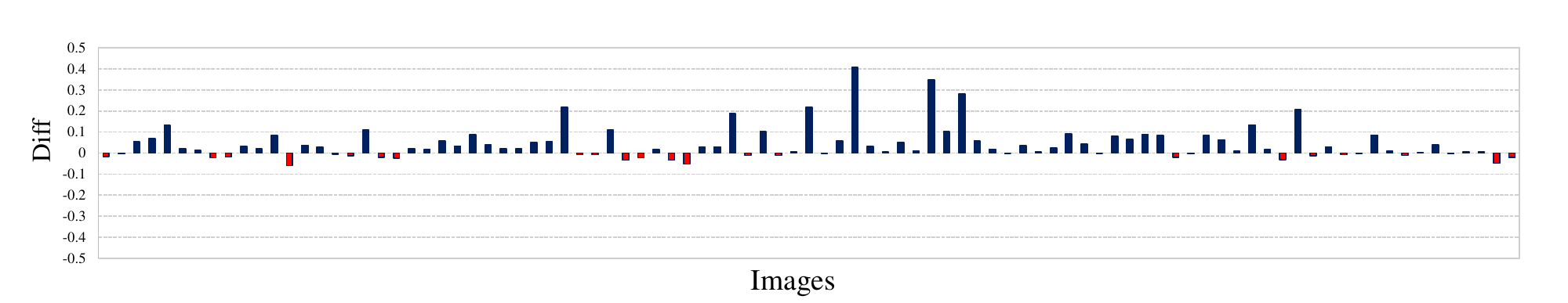}
    \vspace{-5pt}
    \caption{The difference in KL divergence on the DRealSR test set. We can see that our method effectively reduces $D_{kl}$.}
    \label{fig:visfeat_further}
    \vspace{-5pt}
\end{figure*}

As shown in Figure~\ref{fig:visfeat_further}, we calculate the difference in KL divergence $\operatorname{Diff}$ on the DRealSR test set. Define the LR image as $\mathbf{I}$, the formula we use to calculate $\operatorname{Diff}$ is as follows:
\begin{equation}
\operatorname{Diff}(\mathbf{I})= D_{kl(ControlNet)} - D_{kl(Ours)},
\end{equation}
where $D_{kl(ControlNet)}$ is the KL divergence between the control signal of ControlNet and the latent LR embeddings of $\mathbf{I}$, the $D_{kl(Ours)}$ is the KL divergence between the control signal of our model and the latent LR embeddings of $\mathbf{I}$. One can observe that in Figure~\ref{fig:visfeat_further}, the $\operatorname{Diff}$ values for most images are greater than 0, proving that our method can effectively reduce $D_{kl}$. This result indicates that our method can effectively use latent LR embeddings to constrain ControlNet in the latent space.

\begin{figure*}[htp!]
    \centering
    \small
    \setlength{\abovecaptionskip}{0pt}
    \scalebox{0.97}{
    \begin{overpic}[width=\linewidth]{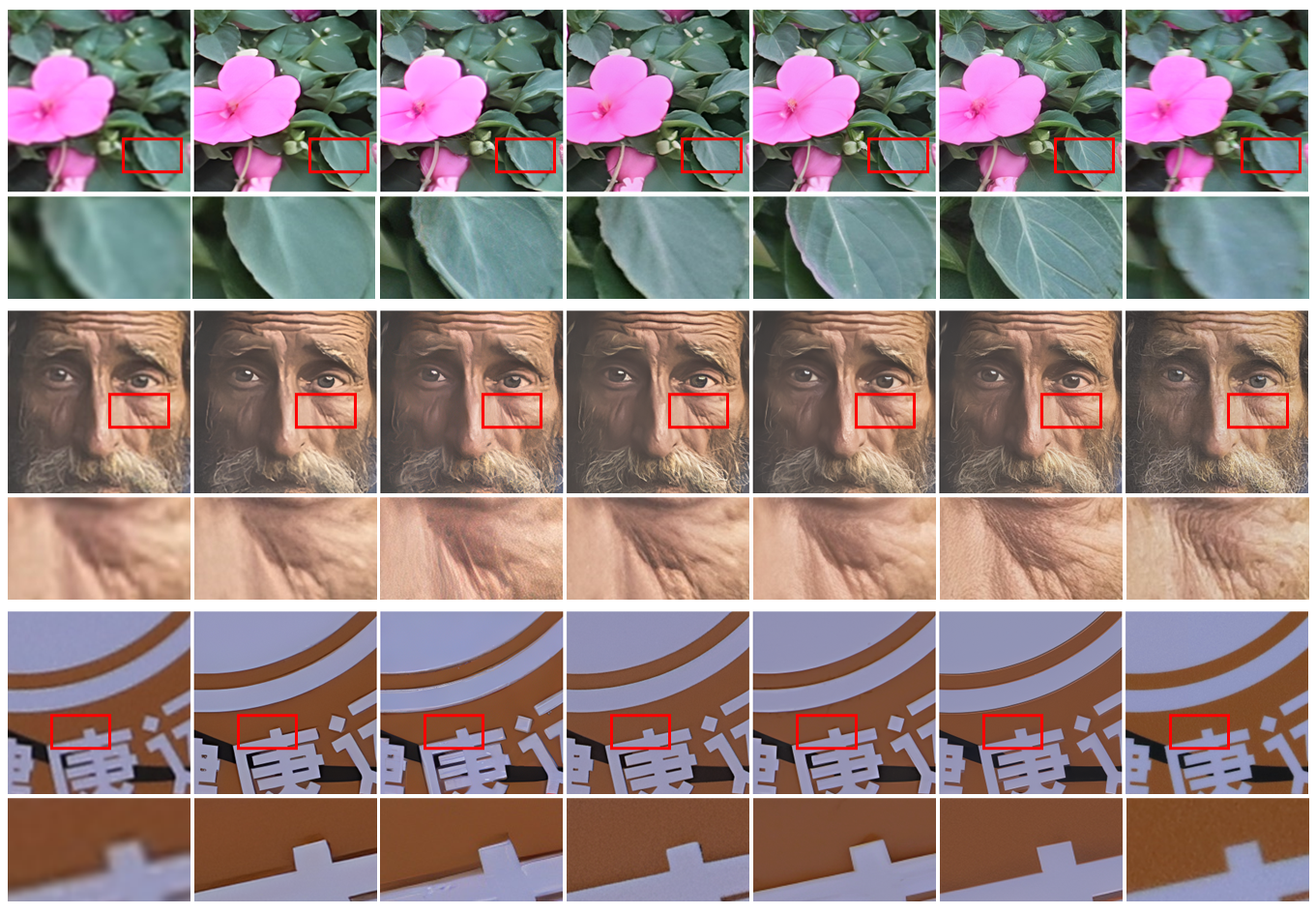}
    \end{overpic}
    }
    \scalebox{0.968}{
    \begin{overpic}[width=\linewidth]{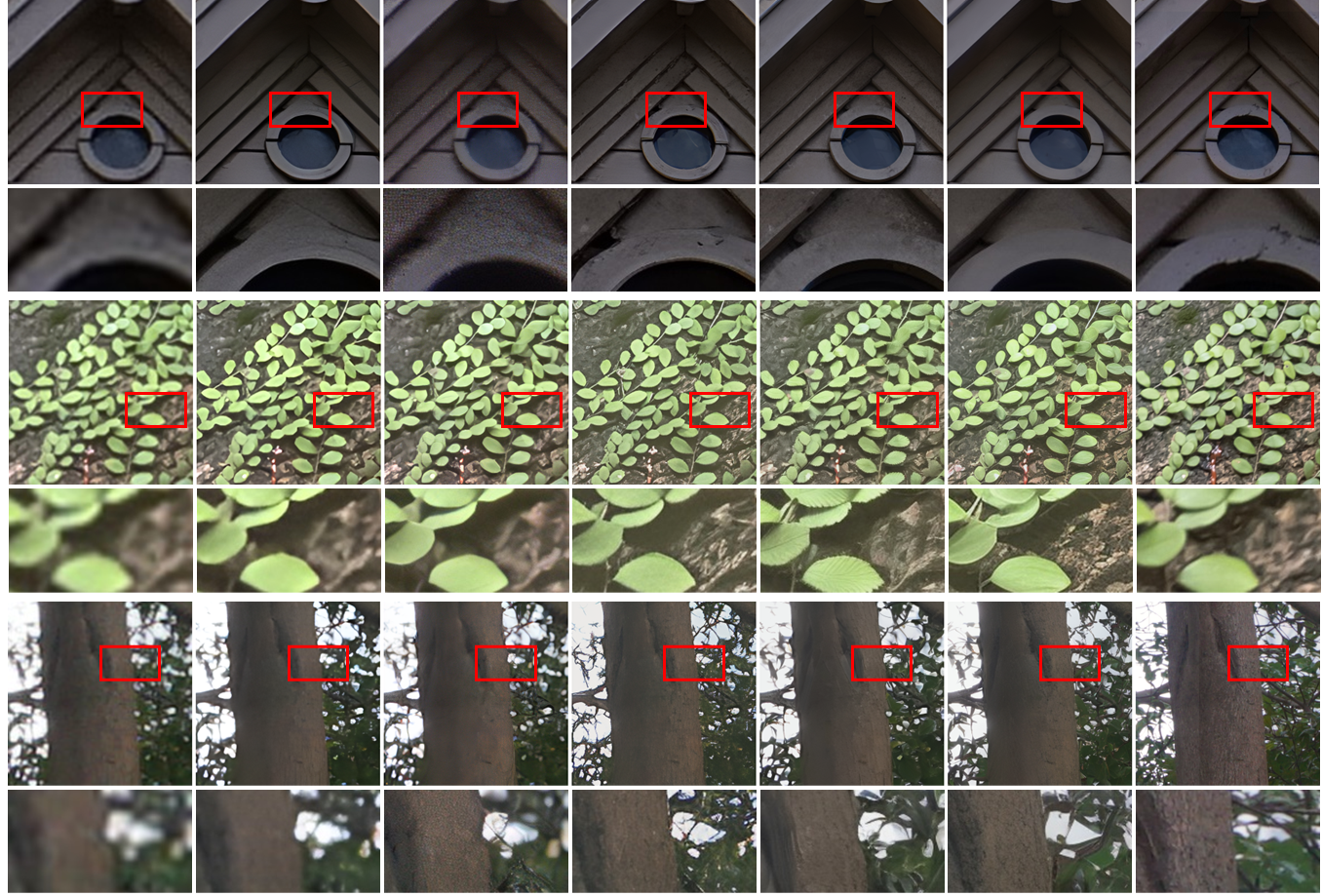}
    \put(3.2, -2){\footnotesize\textcolor{black}{Zoomed LR}}
    \put(16.2, -2){\footnotesize\textcolor{black}{Real-ESRGAN}}
    \put(33, -2){\footnotesize\textcolor{black}{ResShift}}
    \put(46.7, -2){\footnotesize\textcolor{black}{StableSR}}
    \put(61.8, -2){\footnotesize\textcolor{black}{SeeSR}}
    \put(71.5, -2){\footnotesize\textcolor{black}{ControlSR (Ours)}}
    \put(91.2, -2){\footnotesize\textcolor{black}{HR}}
    \end{overpic}
    }
    \vspace{12pt}
    \caption{Visual comparisons with recent state-of-the-art Real-ISR methods. We can see that the results of our \nameofmethod{} have more generated details, and are more consistent with LR images (Zoom in for a better view).}
    \label{fig:visual_compare}
\end{figure*}

\subsection{Comparisons with State-of-the-Art Methods}
\myPara{Quantitative comparisons.}
We show the quantitative comparisons between our \nameofmethod{} and previous state-of-the-art Real-ISR methods~\cite{Real-ESRGAN, LDL, ResShift, SinSR, StableSR, PASD, DiffBIR, SeeSR} in Table~\ref{tab:performance_sd+diffusion}. As GAN-based and Diffusion-based (no generative priors) methods (Real-ESRGAN, LDL, ResShift, SinSR) focus more on fidelity, while SD-based methods focus more on generation, we show the standard \nameofmethod{} to compare with GAN-based and Diffusion-based (no generative priors) methods, and compare another version of our \nameofmethod{} represented as \nameofmethod{*} with modified LSA settings with SD-based methods. As shown in Table~\ref{tab:performance_sd+diffusion}, it can be seen that our method has advantages on all four generation metrics (NIQE, MUSIQ, MANIQA, CLIPIQA) while maintaining high fidelity (PSNR, SSIM).

\myPara{Visual comparisons.}
We show the visual comparisons between our \nameofmethod{} and previous state-of-the-art Real-ISR methods~\cite{Real-ESRGAN, ResShift, StableSR, SeeSR} in the Figure~\ref{fig:visual_compare}. In the first picture, our \nameofmethod{} can generate more realistic leaf vein textures, and in the second picture, our \nameofmethod{} can generate more realistic facial details, demonstrating the superior generative capability of our \nameofmethod{}. Besides, in the third picture, one can observe that our results are clearer than previous methods, proving the effectiveness of our method.

\subsection{Ablation Analysis}
\label{sec:ablation}

In this subsection, we conduct extensive experiments to show the effectiveness of our method. We use DRealSR~\cite{DRealSR} for testing and PSNR, SSIM, NIQE, MUSIQ, MANIQA metrics for evaluation.

\begin{table}
    \centering
    \setlength{\abovecaptionskip}{2pt}
    \setlength\tabcolsep{4pt}
    \renewcommand{\arraystretch}{1.1}
    \small
    \centering
    \caption{
     Ablation on model design.
    }
        \begin{tabular}{lcccc}
    
        \toprule
        {Model Design} & {PSNR$\uparrow$} & {SSIM$\uparrow$} & {NIQE$\downarrow$} & {MANIQA$\uparrow$}  \\
        \midrule
        
         w/o GSPM & 27.57 & 0.7490 & 6.9713 &  0.6241  \\
         DPM w/o cross-attn layers & 28.06 & 0.7358 & 6.1502 & 0.6100  \\
         DPM w/o window partition & 27.60 & 0.7420 & 6.0362 & 0.6273  \\
         full model & 27.93  & 0.7455 & 6.2031 & 0.6219 \\
        
        \bottomrule
    \end{tabular}
    
    \label{tab:ab_model_design}
\end{table}

\begin{table}
    \centering
    \setlength\tabcolsep{4pt}
    \renewcommand{\arraystretch}{1.1}
    \small
    \centering
    \caption{
      Ablation on LoRA layers.}
    \vspace{0pt}
    \begin{tabular}{cccccc}
        \toprule
        {VAE LoRA} & {UNet LoRA} & {PSNR$\uparrow$} & {SSIM$\uparrow$} & {NIQE$\downarrow$} & {MANIQA$\uparrow$} \\
        \midrule
        - & - & 28.89 & 0.7978 & 7.6531 & 0.5381   \\
        $\checkmark$ & - & 27.38  & 0.7296 & 6.2722 & 0.6374   \\
        - & $\checkmark$ & 28.87 & 0.7920 & 7.4344 & 0.5612  \\
        $\checkmark$ & $\checkmark$ & 27.93  & 0.7455 & 6.2031 & 0.6219  \\
    
        \bottomrule
    \end{tabular}
    \label{tab:ab_lora}
\end{table}

\myPara{Effectiveness of Model Design.}
We first conduct the ablation study on our model design. As shown in Table~\ref{tab:ab_model_design}, we compare the full model with several modified versions. As discussed above, GSPM preserves the structural information and DPM preserves the detailed information. We can see the model without GSPM results in a drop in PSNR, which means a loss of fidelity. Moreover, removing the extra cross-attention layers in DPM leads to a drop in MANIQA, which means a loss for generation. We also testing the window partition strategy. We can see that not using the window partition strategy in DPM weakens the fidelity (PSNR and SSIM metrics). These results prove the effectiveness of our model design. 

\begin{table}[t]
\caption{
Ablation on the VAE LoRA rank.
}
\vspace{-6pt}
\label{tab:ab_vae_lora_rank}
\setlength\tabcolsep{4pt}
\renewcommand{\arraystretch}{1.1}
 \small
\centering
    \begin{tabular}{ccccc}
        \toprule
        {VAE LoRA rank} & {PSNR$\uparrow$} & {SSIM$\uparrow$} & {MUSIQ$\uparrow$} & {MANIQA$\uparrow$}   \\
        \midrule
        8 & 27.70 & 0.7512 & 66.64 & 0.6221   \\
        16  & 27.62  & 0.7483 & 66.80 & 0.6222  \\
        32 & 27.46 & 0.7346 & 67.06 & 0.6311  \\
    
        \bottomrule
    \end{tabular}
\end{table}

\begin{table}[t]
\caption{
Ablation on the UNet LoRA rank.
}
\vspace{-6pt}
\label{tab:ab_unet_lora_rank}
\setlength\tabcolsep{4pt}
\renewcommand{\arraystretch}{1.1}
 \small
\centering
    \begin{tabular}{ccccc}
        \toprule
        {VAE LoRA rank} & {PSNR$\uparrow$} & {SSIM$\uparrow$} & {MUSIQ$\uparrow$} & {MANIQA$\uparrow$}   \\
        \midrule
        8 & 27.70 & 0.7508 & 66.28 & 0.6165   \\
        16  & 27.62  & 0.7483 & 66.80 & 0.6222  \\
        32 & 27.83 & 0.7533 & 66.35 & 0.6166  \\
    
        \bottomrule
    \end{tabular}
\end{table}

\begin{table}
\centering
\caption{
Ablation on the Latent Space Adjustment (LSA) strategy. We can see that the LSA strategy can improve fidelity and generation simultaneously.
}
\label{tab:ab_infer}
\vspace{5pt} 
\setlength\tabcolsep{8pt}
\renewcommand{\arraystretch}{1.1}
\small
\begin{tabular}{ccccccc}
        \toprule
        {ELA ${\alpha}=0.01$} & {LLA ${\beta}=0.01$} & {PSNR$\uparrow$} & {SSIM$\uparrow$} & {NIQE$\downarrow$} & {MANIQA$\uparrow$}   \\
        \midrule
        - & - & 28.11  & 0.7419 & 6.5289 & 0.6226   \\
        $\checkmark$ & - & 28.44  & 0.7609 & 6.4765 & 0.6172  \\
        - & $\checkmark$ & 27.85 & 0.7412 & 5.9875 & 0.6360  \\
        $\checkmark$ & $\checkmark$ & 28.22 & 0.7538 & 6.0867 & 0.6246  \\
    
        \bottomrule
     \end{tabular}
\end{table}

\begin{figure*}[!t]
    \centering
    \small
    \setlength{\abovecaptionskip}{0pt}
    \includegraphics[width=\linewidth]{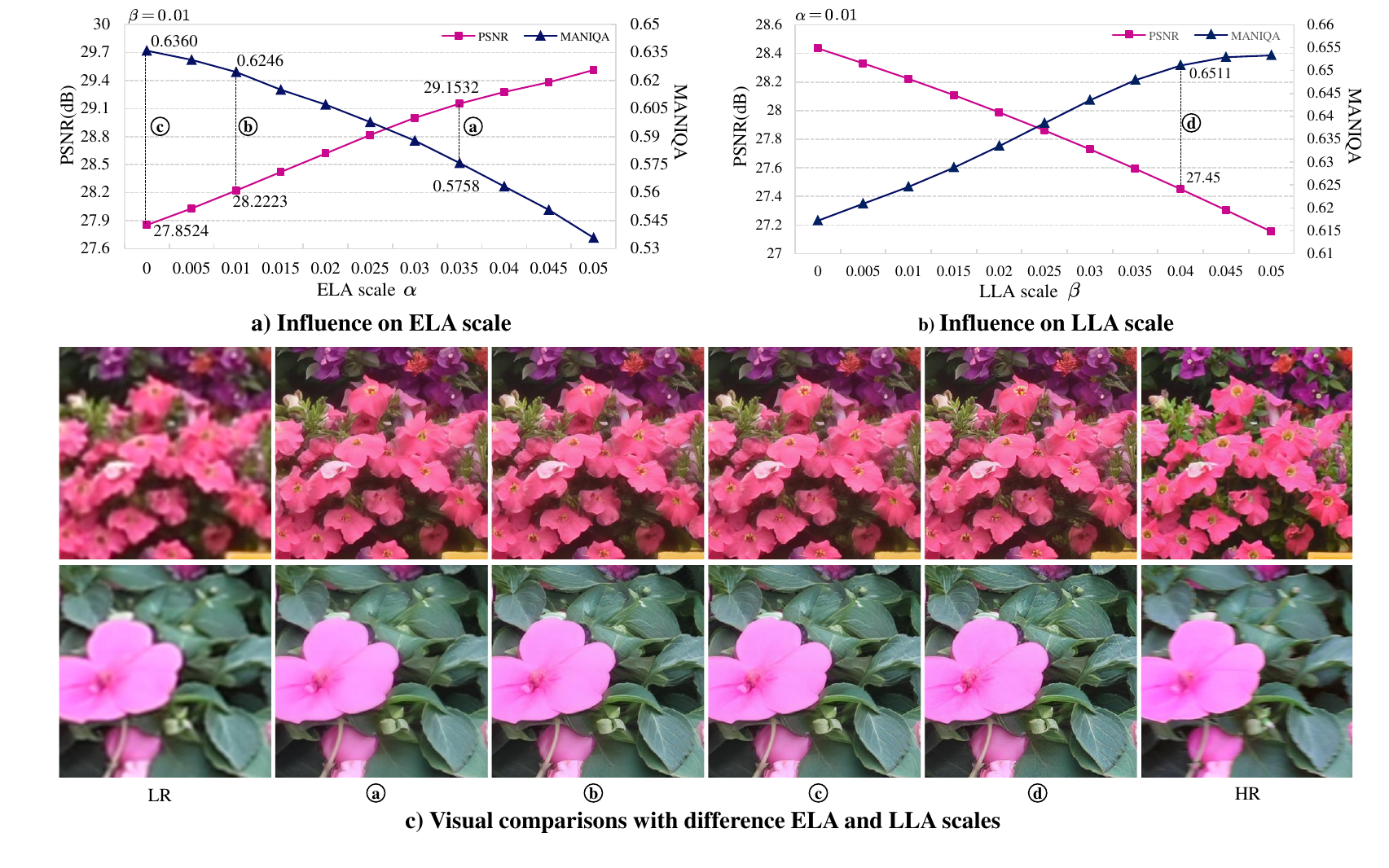}
    \vspace{-5pt}
    \caption{Impact of the Latent Space Adjustment (LSA) strategy in inference stage. a) and b) show the changes in metrics under different settings. c) shows the results under different LSA settings. One can observe that the super-resolution results can be adjusted over a wide range (over 2dB in PSNR and 0.1 in MANIQA).}
    \label{fig:infer_ab}
\end{figure*}

\myPara{Effectiveness of LoRA.}
Next, we demonstrate the effectiveness of LR information adaptation. As shown in Table~\ref{tab:ab_lora}, the model without VAE LoRA layers achieves a very high PSNR and SSIM, but its generative capability decreases significantly. This is because the pre-trained VAE encoder cannot correctly map the LR image to the latent space. The model with VAE LoRA layers and without UNet LoRA layers exhibits higher MANIQA but lower fidelity (PSNR and SSIM metrics). This may be due to the fact that the UNet fails to adapt to the output of the mixed control signals, leading to the misapplication of the provided structural information in generating details.

\myPara{Ablation on the LoRA rank.} We also conduct the ablation study on the LoRA rank. Table~\ref{tab:ab_vae_lora_rank} shows the ablation results for the VAE LoRA rank. As seen, reducing LoRA rank improves fidelity but has a negative impact on generation metrics. On the contrary, increasing LoRA rank has a negative impact on fidelity but improves generation metrics. To balance fidelity and generation, we finally set the VAE LoRA rank to 16. Table~\ref{tab:ab_unet_lora_rank} shows the ablation results for the UNet LoRA rank. We can see that both smaller LoRA rank and larger LoRA rank improve fidelity but have a negative impact on generation metrics. To balance fidelity and generation, we finally set the UNet LoRA rank to 16.

\myPara{Effectiveness of latent space adjustment.}
We demonstrate the effectiveness of our Latent Space Adjustment (LSA) strategy here. Table~\ref{tab:ab_infer} shows the effectiveness of LSA on our \nameofmethod{}. We can see that with appropriate LSA settings, the fidelity and generation of the model can be improved simultaneously. Furthermore, as shown in Figures~\ref{fig:infer_ab}(a) and (b), increasing $\alpha$ can improve the fidelity (See PSNR score), while increasing $\beta$ can improve the generation ability (See MANIQA score).  By using the LSA strategy, the super-resolution results can be adjusted over a wide range (over 2dB in PSNR and 0.1 in MANIQA). Figure~\ref{fig:infer_ab}(c) shows the visual comparisons with difference $\alpha$ and $\beta$. We can see that our \nameofmethod{} can take into account both fidelity and generation. When the PSNR score is high, our model can still generate meaningful textures instead of overly smooth results.

\section{Conclusions}
We propose ControlSR, a new method that can tame Diffusion Models for consistent Real-ISR. We impose stronger constraints on the control signals in latent space through latent LR embeddings, and propose DPM and GSPM to extract LR information at the detail and structure levels. We propose a new LSA strategy in inference stage, which can improve the fidelity and generation ability simultaneously. Extensive experimental results show that the super-resolution results of our ControlSR are more consistent with the LR images and can also generate rich details for better visual effects.

\section*{Data Availibility}
Our method is based on Stable Diffusion 2.1-base~\cite{rombach2022high}, which is publicly available. Our \nameofmethod{} is trained on DIV2K~\cite{DIV2K}, Flickr2K~\cite{Flickr2K}, DIV8K~\cite{DIV8K}, OST~\cite{OST}, and the first 10K face images from FFHQ~\cite{FFHQ}. We use the degradation pipeline of Real-ESRGAN~\cite{Real-ESRGAN} to obtain LR/HR pairs. All these datasets and the degradation pipeline are publicly available.

\bibliographystyle{splncs04}
\bibliography{IEEEabrv,mybibfile}

\end{document}